**Selecting Interpretability Techniques for Healthcare Machine Learning models**

Daniel Sierra-Botero[1,2], Ana Molina-Taborda[1,2], Mario S. Valdés-Tresanco[1], Alejandro Hernández-Arango[3,4], Leonardo Espinosa-Leal[5], Alexander Karpenko, Olga Lopez-Acevedo[1,2]*

1. Biophysics of Tropical Diseases Max Planck Tandem Group, University of Antioquia UdeA, 050010 Medellin, Colombia
2. Grupo de Física Atómica y Molecular, Facultad de Ciencias Exactas y Naturales, Universidad de Antioquia UdeA, 050010 Medellin, Colombia
3. Department of Internal Medicine, University of Antioquia, School of Medicine Medellin, Colombia
4. Hospital Alma Máter de Antioquia, University of Antioquia. Medellin, Colombia
5. Graduate School and Research, Arcada University of Applied Sciences, Jan-Magnus Janssonin aukio 1, 00560, Helsinki, Finland

*Corresponding Author. Email: olga.lopeza@udea.edu.co

**Resumen:** En el campo de la atención médica, se busca emplear algoritmos interpretables para ayudar a los profesionales de la salud en diversos escenarios de toma de decisiones. Siguiendo el marco Predictivo, Descriptivo y Relevante (PDR), se define el aprendizaje automático interpretable como un modelo de aprendizaje automático que de manera explícita y simple expone las relaciones determinadas en los datos o aprendidas por el modelo que son relevantes para su funcionamiento, y se categorizan los modelos como post hoc, adquiriendo interpretabilidad después del entrenamiento, o basados en el modelo, que están intrínsecamente integrados en el diseño del algoritmo. En este articulo presentamos una selección de ocho algoritmos, tanto post hoc como basados en el modelo, que pueden utilizarse para estos fines.

**Palabras claves:** Aprendizaje de máquinas, Inteligencia Artificial, Interpretabilidad

**Abstract:** In healthcare there is a pursuit for employing interpretable algorithms to assist healthcare professionals in several decision scenarios. Following the Predictive, Descriptive and Relevant (PDR) framework, the definition of interpretable machine learning as a machine-learning model that explicitly and in a simple frame determines relationships either contained in data or learned by the model that are relevant for its functioning and the categorization of models by post-hoc, acquiring interpretability after training, or model-based, being intrinsically embedded in the algorithm design. We overview a selection of eight algorithms, both post-hoc and model-based, that can be used for such purposes.

**Keywords:** Machine Learning, Artificial Intelligence, Interpretability

As the capabilities of machine learning models evolve, so does the urgency to understand the decision-making processes within them. This pursuit has led to an array of different related terms with various definitions depending on the authors and the context, such as explainability, transparency, intelligibility, etc. However, for the purpose of this article, we will adopt the definition of interpretability provided in the Predictive, Descriptive, and

Relevant (PDR) framework. This framework defines interpretable machine learning as "the extraction of relevant knowledge from a machine-learning model concerning relationships either contained in data or learned by the model" (1). This definition encloses many processes that can be categorized as interpretable and that vary drastically depending on the type of problem that is trying to be solved.

Interpretable algorithms prioritize clarity and understanding over statistical inference. Instead of aiming for causal relationships, they emphasize the identification of correlations, which, while informative, can be less robust. In the context of healthcare decisions, as stated by Rudin and coworkers (2), for high-stakes decisions, interpretability takes on a particular role. Rather than promoting blind trust in algorithmic outputs, it strives to establish a framework that empowers healthcare professionals, such as doctors, to exercise their expertise and judgment. This approach acknowledges that medical decisions require a delicate balance between algorithmic assistance and human judgment, allowing healthcare providers to assess the algorithm's recommendations and determine the appropriate level of trust.

In this article, we want to investigate the intricate domain of interpretability in Machine Learning, aiming to provide insight into the great variety of existing methods and how they work. At the center of our exploration lies two distinct approaches to achieve interpretability. The model-based approach involves decisions made prior to the model's training, with the goal of enhancing its interpretability. These decisions, integrated into the training process, ease a clearer understanding of how the model arrives at its final decisions. On the other hand, the post hoc approach involves techniques applied after the model has been trained. These techniques extract insights from the trained model, shedding light on its inner workings and making it more accessible to human comprehension.

Algorithms reviewed in this article are classified according to their interpretable potential in health care problems and summarized in Table 1. According to a recent checklist type proposal based on the PDR framework, ideal algorithms in healthcare would have interpretability in one or more of the following five characteristics (3): Feature importance, Descriptive accuracy, Simulatability, Relevance and Predictive accuracy. Feature importance collects algorithms that convey the importance of feature in determining algorithm outputs for a given prediction. In the PDR framework they are classified then as post-hoc approaches. We will review some perturbation-based and gradient-based methods as examples. Following the model-based approach, in literature also denoted as the intrinsic approach, the algorithms that seek a Descriptive accuracy do so by designing the algorithm a-priori to have the most interpretable structure. Simple models that can be easily understood by examining their mathematical structure have higher descriptive accuracy compared to more complex models where the mathematical formula is too intricate for human comprehension (black box models) (4). In the PDR frame they use as example a comparison of simple ML models vs deep networks in classification of severity of illness which leads to the same accuracy (5,6). Apart from choosing the right complexity, we exemplify this category here with the Generalized and Scalable Optimal Sparse Decision Trees algorithm (GOSTD). Thanks to their complex structure, black box models often excel at capturing nonlinear relationships present in the data and making accurate predictions. But this can lead to the "Clever Hans" effect, where the model makes the right decisions for the wrong reasons, and the complexity of the model makes it nearly impossible to detect such cases (7). To avoid this instead of merely explaining black box models, we must consider the value of inherently interpretable models.

Simulatability would refer also to model-based algorithms designed so that "clinicians can understand and mentally simulate the model's process for generating predictions" (3). We present here risk score algorithms as an example to use with this end. Relevance describes "relevancy as judged by the algorithm's target human audience" (3) and we present the algorithms ProtoPNet (This Looks Like That) and interactive reconstruction as examples. Finally, Predictive accuracy would refer to algorithms that quantify accuracy inside a given class of a patient. Then if the class has a high accuracy error the doctor would be able to deemphasize the algorithm use. As example of such a class of algorithms we include here influence functions.

Classifying interpretability into local and global methods helps users selecting appropriate interpretability techniques. Local methods explain specific small parts of the model's prediction space, while global methods offer an overall understanding of the model but may struggle to explain individual predictions.

**1. Local Interpretable Model-agnostic Explanations (LIME)**

Feature importance conveys the relative importance of features in determining algorithm outputs. A feature is an individual measurable property or characteristic of a phenomenon, such as blood pressure, heart rate, or age, that can be used by a machine learning algorithm to learn patterns or make predictions. In general, the methods used for determining it are called Attribution Methods. These assign an attribution value that quantitatively represents the importance of each feature in the input. The representation of all the attributions of the features in the same form as the input sample are called attribution maps. These attribution maps, having the same dimension as the input, can be superimposed with it, allowing the visualization of the relative importance of each feature of the input to obtain the output.

Although there are different criteria, attribution methods can be divided into two fundamental categories: perturbation-based and gradient-based methods.

"Perturbation-based methods directly compute the attribution of an input feature (or set of features) by removing, masking or altering them, and running a forward pass on the new input, measuring the difference with the original output" (8).

Gradient-based methods use the calculation of gradients of the model outputs with respect to the input features. Gradients are indicators of how the model outputs change relative to changes in the inputs.

Among the perturbation-based methods we can find the additive feature attribution methods. While not only reliant on perturbation, their methodology shares some key elements and definitions with it. These techniques stand out for their ability to illuminate the complex decision-making processes of models, particularly those that appear inscrutable as 'black boxes.' Two standout examples within this category include Local Interpretable Model-agnostic Explanations (LIME) and SHapley Additive exPlanations (SHAP).

LIME takes a unique approach to model interpretability by constructing local interpretations. When a model's prediction for a specific instance needs explanation,

LIME generates an interpretable model, like a linear regression, that approximates the complex model's behavior around one instance. To do this, LIME perturbs the instance by making small changes to the feature values and observes how these changes Rinfluence the model's prediction. By doing so across a range of perturbations, LIME's interpretable model captures the intricate relationship between features and predictions for that instance (9). This is similar to how sensitive analysis in epidemiology evaluates how the uncertainty in the input parameters of a model affects the output to assess the impact, effect or influence of key assumptions or variations (10).

## 2. SHapley Additive exPlanations (SHAP)

On the other hand, SHAP employs a different methodology based on principles of cooperative game theory, attributing values to individual features by considering all possible ways features can interact like how players can interact in coalitional games. The key concept is 'Shapley values', which provide a consistent way to fairly distribute the contribution for each feature across different predictions. The beauty of this approach lies in its ability to reveal how each feature's presence or absence in the model influences predictions, thus offering a clearer picture of the model's decision-making process (11).

This method has been employed to 'explain' mortality risk predictions in patients with nonmetastatic prostate cancer. The study harnessed a set of predefined features, including age, prostate-specific antigen (PSA), Gleason score, etc. And trained a gradient-boosted tree model. When applying SHAP values to explain this model, intriguing insights emerged. By exploring the interaction between PPC and Gleason score, the study found that the interaction effects were notably stronger in patients with

Gleason ≥ 8 disease compared to those with Gleason 6-7 disease, especially when PPC was ≥ 50%. Confirmatory linear analyses further validated this finding (12). This practical application underscores how SHAP values can unravel complex interactions within medical data, offering clinicians a clearer understanding of factors influencing patient outcomes.

However, it is essential to acknowledge certain limitations and potential challenges associated with this technique. While SHAP have desirable mathematical properties and applicability to specific explanations, mathematical complexities can arise when Shapley values are employed for feature importance. The exact computation of Shapley values requires knowledge of an extensive set of multivariate distributions, often leading to the need for feature reduction. Moreover, the choice of which features to include in the analysis becomes critical, as explanations may vary based on the features considered. This selection impacts the resulting attributions, potentially biasing the perceived importance of certain features. Additionally, in some cases, the model may have to make predictions in areas where it has not seen data before, which can lead to unexpected results. This can be a problem if someone tries to manipulate the model (13).

### 3. Gradient-weighted Class Activation Mapping

Gradient-based methods require a single forward and backward pass through the network to produce the attribution map. in simpler terms, the model looks at the input data and calculates how important each feature is for the final prediction and produce a guide called attribution map Additionally, the number of steps in the network does not depend on the number of input features, which allows for easier scaling. On the other hand, they are also easy to implement, and attributions can be calculated for any network architecture or neural network structure (14).

Although gradient-based methods have several advantages, they also have limitations that must be carefully considered. Gradient-based methods are strongly affected by noisy gradients (8,14,15). This leads to attribution maps that can show irrelevant feature contributions. Additionally, as gradient-based methods are local interpretability methods, the explanation provided for a single test sample can lead to erroneous conclusions about the model's performance (16). Also, various gradient-based methods can be manipulated, identifying specific or unimportant tokens as highly important in several Natural Language Processing (NLP) tasks (15). Although several modifications to the gradient-based approach have been proposed to address the challenge of noise in attribution maps (17), gradient-based methods still do not provide a reliable metric in all cases. In this sense, some authors recommend using multiple analysis techniques to guarantee more trustworthy interpretations (15,18).

The most common Artificial Intelligence (AI) methods in medicine are dedicated to image processing. Due to the characteristics of the models applied one of the most widely used post-hoc interpretability methods are Gradient-weighted Class Activation Mapping (Grad-CAM) (19). Grad-CAM is a generalization of Class Activation Mapping (CAM) that does not require a particular architecture and can be computed on any network with gradients. This method allows us to understand which parts of an input image are important for a classification decision by exploiting the spatial information preserved across convolutional layers of the network (20).

Grad-CAM has been applied in numerous research areas and is particularly popular in models related to medical image processing (19). It has been the most widely used interpretability method in non-exhaustive MRI studies using AI and Explainable Artificial Intelligence (XAI) techniques since 2017, with 34% above CAM (30%) and other methods such as LIME (5%) and SHAP (2%) (19). Grad-CAM has been used to validate models generated for the detection of Attention Deficit Hyperactivity and Conduct Disorder (21), detection of different types of cancer (22–34), detection of osteoarthritis (35), cardiac problems (36) and Covid-19 (37–41), just to mention a few examples.

**4. Generalized and Scalable Optimal Sparse Decision Trees algorithm (GOSTD)**

In the context of model interpretation, descriptive accuracy refers to the extent to which an interpretation method objectively captures the relationships learned by the models (1). High descriptive accuracy ML models can be achieved by training them to be inherently interpretable from the start, constraining the model's domain and minimization in a way that makes them more understandable to humans. These constraints not only shed light on the model itself but also enhance our ability to comprehend its predictions (4).

The primary challenge in constructing such models lies in the process of minimizing the loss function while adhering to the constraints, which can be computationally expensive. Furthermore, the mathematical solution for this optimization varies significantly depending on the type of data, the problem to solve and the scoring metric used. The Generalized and Scalable Optimal Sparse Decision Trees (GOSDT) algorithm is a novel approach to decision tree optimization that is designed to handle a wide variety of scoring functions, including weighted accuracy, F-score, AUC, and partial area under the ROC curve. Unlike previous methods like Certifiable Optimal RulE ListS (CORELS) and OSDT, which are designed solely to maximize accuracy, the GOSDT method employs hash trees to represent similar structures using shared subtrees. This allows for more efficient optimization of decision trees, which can be particularly useful in medical applications where interpretability is important (42).

**5. Risk score**

Following the idea of constructing models with inherent constraints to enhance interpretability, we venture into the realm of simulatability, a concept denoting a model's capacity to be understood and replicated by humans (1). An illustrative example of such models are risk scores. These models are essentially classification models that empower users to assess risk through simple arithmetic operations like addition, subtraction, and multiplication, relying on a handful of easily comprehensible numerical values. They find extensive applications in various decision-making domains, including criminal justice, finance, and notably in healthcare. For instance, they are utilized to gauge outcomes such as mortality risk (43), predict critical physical conditions (44,45), and even aid in diagnosing mental illnesses (46,47).

What sets risk scores apart from traditional scoring systems is their score conversion into kess probabilities. While numerous scoring systems have been devised, especially within the healthcare sector, it's worth noting that most of these systems have not undergone pure algorithmic optimization applied to data. Instead, they have typically been constructed using a heuristic approach, often relying on domain expert opinions backed up by empirical data (48).

Striking the delicate balance of optimizing predictive performance, sparsity, and other user-defined constraints can prove to be a complex undertaking. One notable machine learning method designed to obtain optimal risk scores is RiskSLIM, presenting a promising avenue for the development of interpretable risk scores. This innovative approach addresses the challenges of calibration, sparsity, and coefficient restrictions through a single-shot procedure, offering a more effective solution compared to traditional methods. RiskSLIM's ability to optimize feature selection and obtain small integer coefficients, while accommodating domain-specific constraints without extensive parameter tuning, marks a significant advancement in the quest for interpretable machine learning models.

In an evaluation RiskSLIM demonstrate a superiority over conventional approaches. The consistently better calibration and AUC achieved by RiskSLIM underscore its potential as a powerful tool for creating risk scores that strike an ideal balance between interpretability and predictive performance. As the field of interpretable machine learning continues to evolve, RiskSLIM stands as a notable example of how advanced optimization techniques can pave the way for more transparent and actionable risk assessment tools (48).

## 6. ProtoPNet (This Looks Like That)

In the brain, visual information enters through the retina and then travels through a pathway called the ventral visual pathway. This pathway consists of different areas in the visual cortex such as V1, V2, and V4, and the inferior temporal cortex (ITC). After the initial processing of visual information, which takes about 150 milliseconds, the brain starts to process the information more deeply by activating recurrent processes that connect higher and lower areas of this pathway. This allows the brain to recognize objects and shapes more accurately (49).

ProtoPNet is a model-based approach which main objective is to "identify several parts of the image where it thinks that this part of the image looks like that prototypical part of some class and makes its prediction based on a weighted combination of the similarity scores between parts of the image and the learned prototypes" (50).

ProtoPNet is a deep learning model that can help classify images. It works by breaking down an image into smaller parts called prototypes, and then combining evidence from these prototypes to make a final classification. The model uses a combination of a Convolutional Neural Network (CNN), a Prototypical Layer, and a Fully Connected Neural Network to achieve this.

However, before the model can be used, there are two main preprocessing steps that need to be done. The first step is offline data augmentation, which involves creating more training data by applying various transformations to the original images. The second step is pretraining the model with crop images of each prototype.

It's important to note that this network uses labels for each prototype, but it's unknown for the model what prototype it's referring to. The cost function of the model consists of penalizing the misclassification of the training data, minimizing the difference between patches of the images with close prototypes, and maximizing the difference for not proper prototypes.

As example It was used as a core idea in a medical application for identifying Covid-19 in X-ray scans (51). They proposed a new version of ProtoPNet called Gen-ProtoPNet. The accuracy values of Gen-ProtoPNet and the baseline model were similar, being 87.27% and 88.54%, respectively. However, they found that Gen-ProtoPNet has better performance for this problem than ProtoPNet.

**7. Interactive reconstruction in representation learning**

Generative models are algorithms able to generate new samples of a dataset by learning the probability density distribution of the data. Autoencoders, Variational Autoencoders (VAE) or Generative Adversarial Networks (GANs) are some of the most representative algorithms. Representation learning is one of its applications and refers to find mappings between high-dimensional inputs and low-dimensional representations and the interpretability is focused on finding disentangled representations, however, disentanglement measurements are only applicable to synthetic datasets (52).

Ross et al. (52) proposed a methodology for interactively reconstructing models for synthetic and real-world datasets. However, for this last the ground truth is unknown, therefore the disentanglement representation is not fully reliable. This method assesses the interpretability of generative models by measuring how effectively users can change the representations interactively to recreate the desired target. The reconstruction was performed on the datasets dSprites, Sinelines and MNIST, and for measuring the quality of the reconstructed model they used metrics such as completion rate, response time, slide distance, error AUC and self-reported difficulty for each user. Regarding disentanglement measures, the authors use two disentanglement measures: DCI, disentanglement, completeness, and informativeness score (53) and the mutual information gap (MIG) (54). These measures are used to quantify the extent to which the models match ground truth on the datasets.

They used control of representations, visualization of instances, and defined a choice of distance metric, distance threshold, and time threshold prior to the reconstruction task, and found this task much more reliable when differentiating entangled and disentangled models for synthetic datasets. Some limitations of this method include the assumption of ground-truth knowledge and bias due to visualization of the target instances. They suggest that their proposed method should be used with other methods for evaluating model interpretability, especially in real-world datasets.

## 8. Influence functions

Relevance via Influence functions. The objective of influence functions is to determine the data points in the training data that the model predicts are closer to the data points in the test data and could potentially be used to determine the relevance of the model by the user (55). The distance is measured by the influence that the removal of one training point has on the test loss on that specific point of the test. The total number of model evaluations grows as the number of training points times the number of data points. Therefore, for big data sets, it becomes computationally prohibitive. Some applications include searching for mislabeled data (56) or discarding features that can negatively influence the model prediction (57). Some authors have shown that influence functions are unsuitable as explainable methods in machine learning models based on neural networks regardless of their architecture (58). Despite this, it has been shown its potential for validation in causal inference methods (59).

## Conclusion

In conclusion, our exploration of various techniques encompassing both model-based and post-hoc approaches reveals an open landscape of interpretability tools that can be chosen to address distinct objectives. The presented methods in this work are relevant for healthcare applications, although in the scientific landscape there are more variations or inspired algorithms for specific applications. As reviewed here, no single method is without its shortcomings. Our strong recommendation is that combining these techniques can mitigate individual flaws and result in more robust and comprehensive interpretability solutions (Figure 5). In embracing this diversity of approaches, we ensure that interpretability remains a cornerstone of responsible and trustworthy AI development.

## Acknowledgments

Proyecto Innovación Abierta Hospital Alma Máter de Antioquia y Universidad de Antioquia.

**TABLA 1:** Algorithms overviewed here that are relevant for reaching interpretability in HC applications.

|   | Algorithm | Characteristic or Goal in HC | Approach | Where to find it: web address or packages |
|---|---|---|---|---|
| 1. | Local Interpretable Model-agnostic Explanations (LIME) | Feature Importance | Post-hoc | - https://github.com/marcotcr/lime <br> - https://captum.ai/ |
| 2. | SHapley Additive exPlanations (SHAP) | Feature Importance | Post-hoc | - https://shap.readthedocs.io/en/latest/ <br> - https://captum.ai/ |
| 3. | Gradient-weighted Class Activation Mapping (Grad-CAM) | Feature Importance | Post-hoc | - https://tf-explain.readthedocs.io/en/latest/ <br> - https://captum.ai/ |
| 4. | Generalized and Scalable Optimal Sparse Decision Trees algorithm (GOSTD) | Descriptive Accuracy | Model-based | - https://github.com/Jimmy-Lin/GeneralizedOptimalSparseDecisionTrees |
| 5 | Risk-calibrated | Simulatability | Model- | - https://github.com/ustunb/risk-slim |

| . | Supersparse Linear Integer Model (RiskSLIM) |  | based |  |
|---|---|---|---|---|
| 6. | ProtoPNet (This looks like that) | Relevance | Model-based | - https://github.com/cfchen-duke/ProtoPNet |
| 7. | Interactive reconstruction | Relevance | Post-hoc | -https://github.com/dtak/interactive-reconstruction |
| 8. | Influence functions | Predictive Accuracy | Post-hoc | - https://pypi.org/project/Influenciae/ |

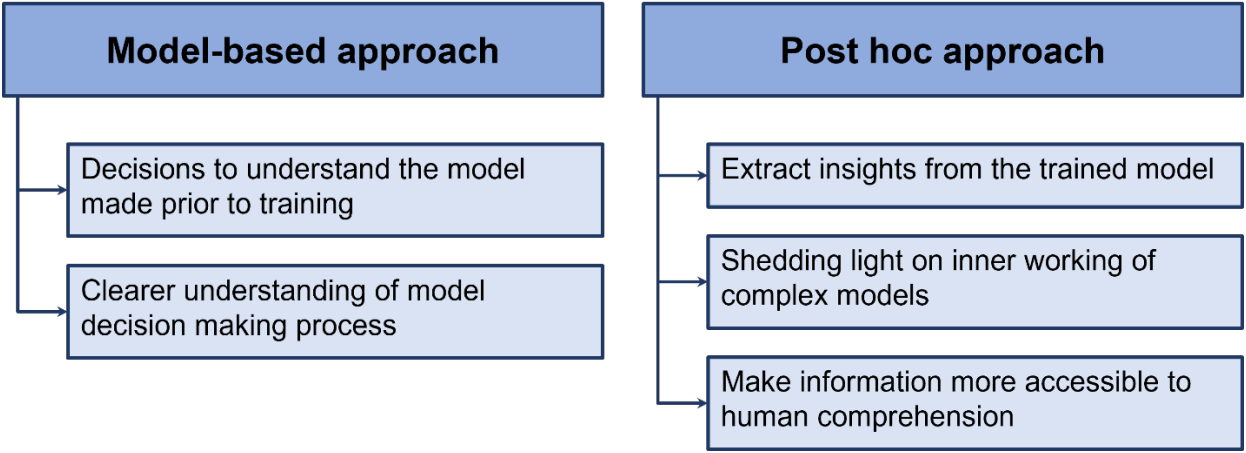

**FIGURA 1**. Categorization of model interpretability techniques based on the training stage in which they are performed.

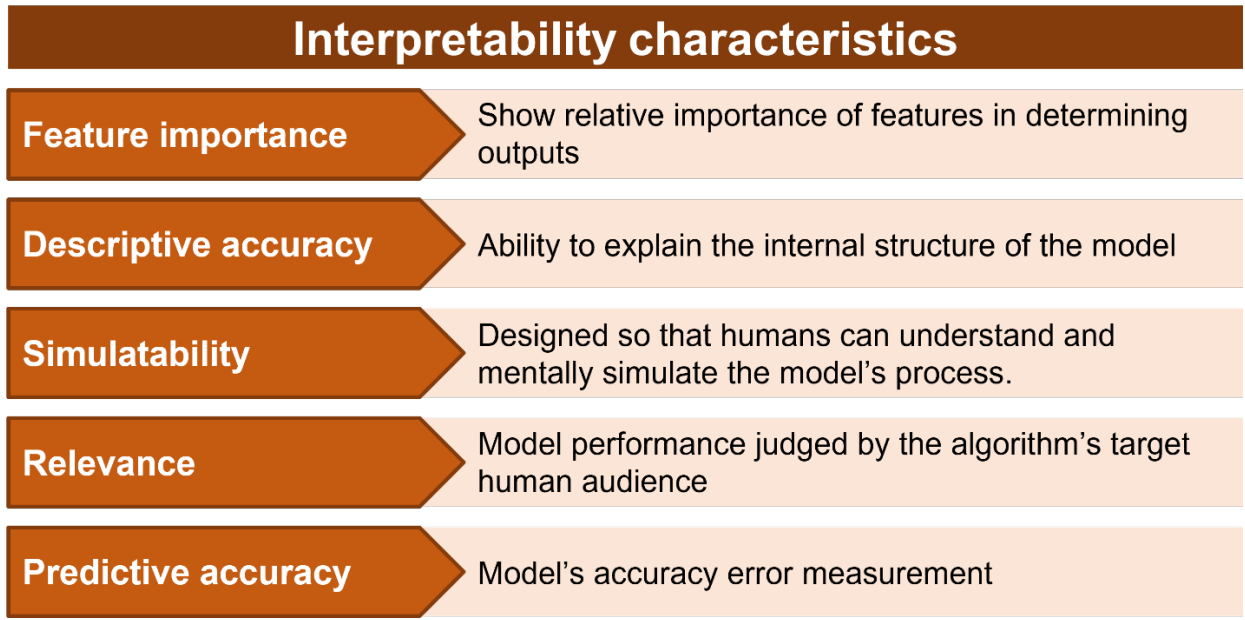

**FIGURA 2.** Interpretability characteristics that an ideal algorithm must meet in healthcare, list adapted from (3).

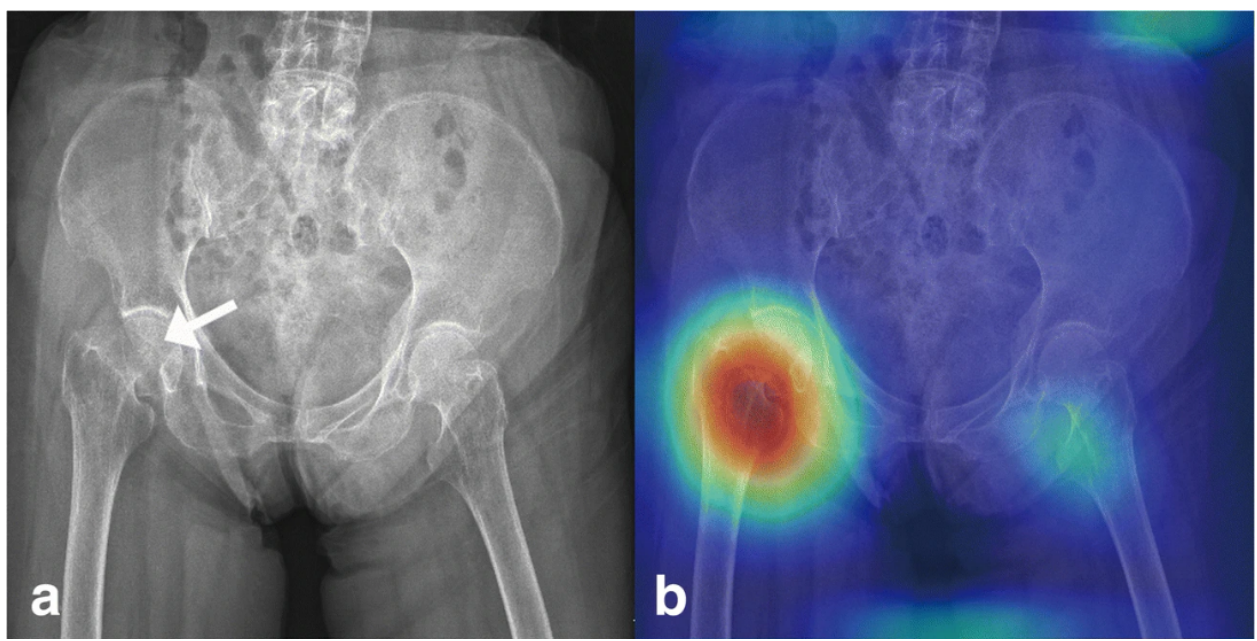

**FIGURA 3.** Grad-CAM heatmap representation over a hip fracture medical image. **a)** the original pelvic radiograph with a mildly displaced right femoral neck fracture (arrow) and **b)** the image generated after applying the model with Grad-CAM, which visualizes the class-discriminative regions, as the fracture site. Image (modified) taken from (50).

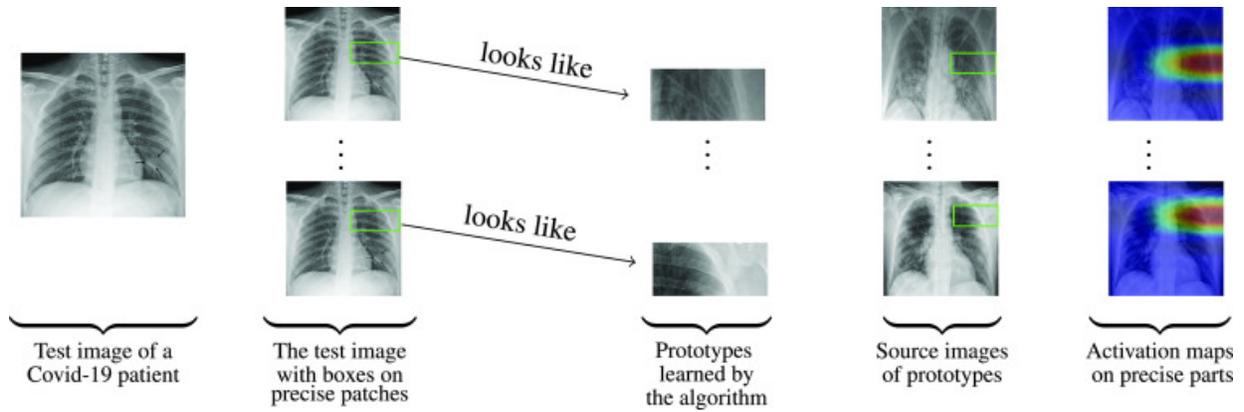

**FIGURA 4**. Visual results of the Gen-ProtoPNet, comparing an image of a Covid-19 patient with two learned prototypical parts and its respective activation map. Image taken from (51).

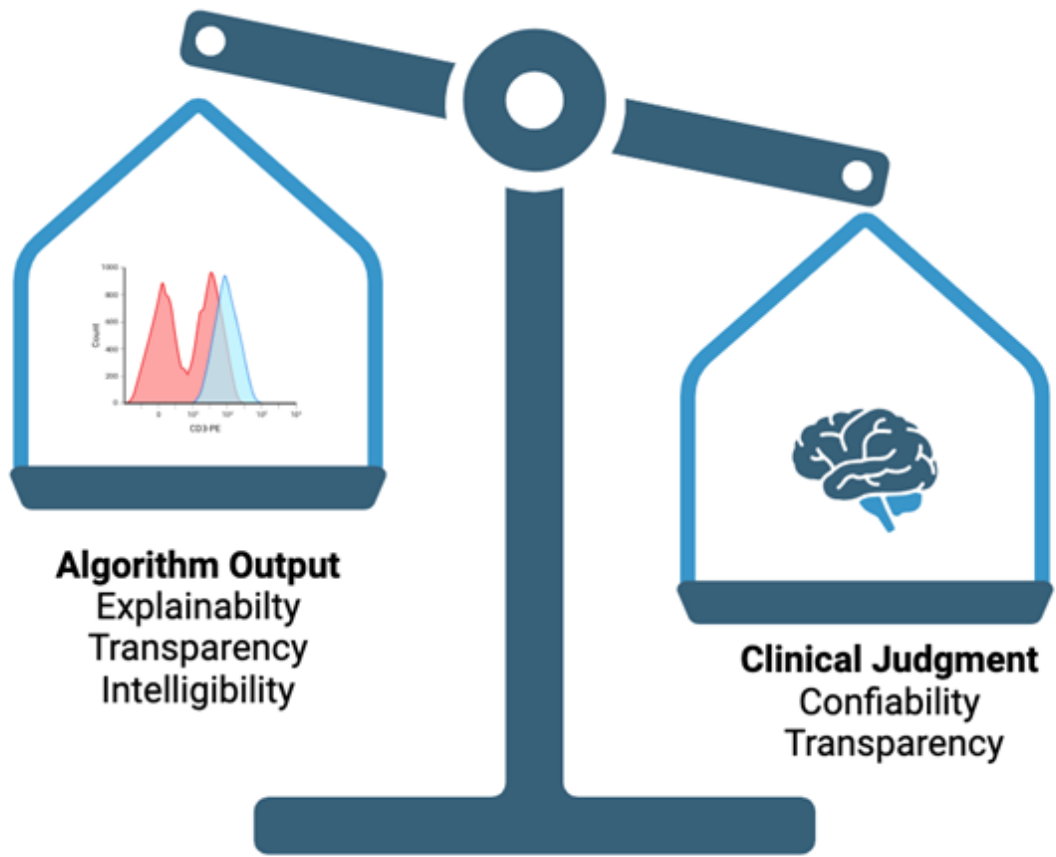

**FIGURE 5**. By conditioning a prediction algorithm to be interpretable, intelligible and transparent, an AI prediction can be assessed/incorporated by clinical reasoning and be a better tool for enhancing clinical judgment.